\documentclass[sigconf,nonacm]{aamas}
\usepackage{amsmath}
\usepackage{booktabs}
\usepackage{balance}
\AtBeginDocument{%
  \fancypagestyle{standardpagestyle}{\fancyhf{}%
    }%
  \fancypagestyle{firstpagestyle}{\fancyhf{}%
    }%
  \pagestyle{standardpagestyle}}

\title{Same Executions, Different Gaps: Judgment Rules in Web-Agent Evaluation}

\author{Zelong Wang}
\affiliation{%
  \institution{University of California San Diego}
  \city{La Jolla}
  \country{USA}}
\email{zew013@ucsd.edu}

\author{Weiyue Li}
\affiliation{%
  \institution{Harvard University}
  \city{Cambridge}
  \country{USA}}
\email{weiyueli@fas.harvard.edu}

\author{Fei Xue}
\affiliation{%
  \institution{Independent Researcher}
  \city{Seattle}
  \country{USA}}
\email{sjtusherffy@gmail.com}

\author{Junru Zhu}
\affiliation{%
  \institution{Independent Researcher}
  \city{Seattle}
  \country{USA}}
\email{junru.zhuu@gmail.com}

\begin{abstract}
Success-rate gains guide the development of web agents. Yet knowing which agent ranks first does not tell us how much confidence to place in its reported advantage. We study the comparison itself: how success is defined, what evidence supports the judgments, and how the difference between agents is estimated. On fixed web-agent executions, changing a judge from predicting likely completion to requiring recorded support can remove most of a reported lead without changing the ranking. This effect persists when unresolved completion is distinguished from supported failure. Controlled source additions explain how missing relations between task requirements, objects and observations can conceal a difference or create an apparent advantage. We also find that making both agents' score estimates more accurate can make their difference less accurate, because their original errors partly canceled. To estimate a chosen outcome difference, we combine judge predictions with reference labels for sampled task pairs. Across four released execution systems, regression fitted on the same 50 labeled pairs reduces estimation error relative to direct paired sampling. The results give agent researchers a practical basis for assessing claimed improvements: make the comparison target explicit, retain the evidence needed to check it, and use paired labels to estimate the difference directly.
\end{abstract}
\begin{document}
\maketitle

\section{Introduction}
A new web agent is usually evaluated through its improvement over a baseline. The size of that improvement helps researchers decide whether a change in planning, perception or tool use is worth pursuing. Model judges make these comparisons easier to scale: they inspect saved executions, assign success labels, and turn those labels into scores. But a plausible judgment of each execution is only part of the evidence needed to support a claim that one agent is better by a particular amount.

The difference between two scores can behave differently from either score alone. If an evaluation rule lowers both scores unequally, the ranking may stay the same while the reported improvement largely disappears. More accurate scores can even yield a worse difference when their original errors partly canceled. The claimed advantage therefore needs its own check. Agreement on a winner, or better accuracy for each agent separately, leaves its magnitude unresolved.

We study this problem on fixed executions from AgentRewardBench~\cite{Lu2025}. A judge can be asked two useful questions: did the task likely succeed, or does the supplied record support completion? Holding executions and records fixed, switching between these questions reduces Claude's lead over Qwen from 8.33 to 1.33 percentage points in one setting. The point-estimate ranking is unchanged. A new comparison that explicitly allows an unresolved verdict retains the reduction. Additional judges and execution systems show a range of responses, making the meaning of an improvement claim part of its evaluation.

Checking that meaning also requires checking the evidence. A product title alone may not establish that an agent clicked the requested grid position; a purchase confirmation may omit which configuration was selected. Our source interventions recover relations needed to assess such conditions. They reveal a missed distinction in one comparison and remove an apparent advantage in another. These opposite outcomes illustrate a common objective: make the judgment traceable to the task and its recorded result.

\begin{figure*}[tp]
\centering
\includegraphics[width=\textwidth]{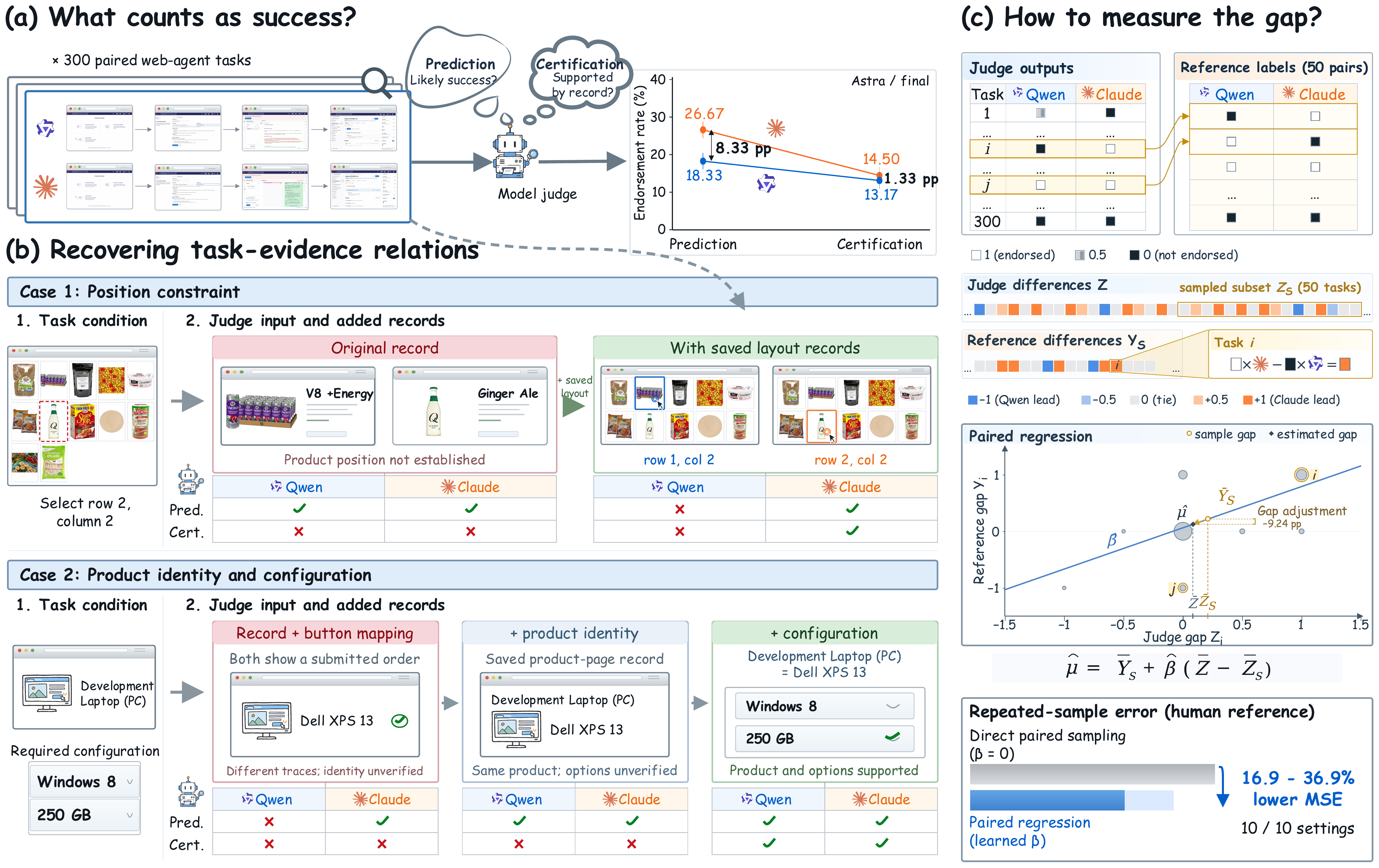}
\caption{Three complementary checks of an agent comparison. \textbf{(a)} Astra judges the same full trace and final screenshot under two success rules. \textbf{(b)} Archived task relations reveal a positional distinction and remove an apparent purchase advantage; browser cards illustrate record contents. Checks denote success endorsement; crosses denote non-endorsement. \textbf{(c)} Full-cohort judge differences and 50 paired reference labels yield a regression-adjusted estimate. A judge value of 0.5 averages two disagreeing calls. Strips show task excerpts; the scatter illustrates one sample, while the bars summarize repeated-sampling MSE across ten human-reference settings.}
\Description{Three complementary studies show rule-dependent endorsement gaps, local effects of adding layout or product records, and paired regression that combines full-cohort judge outputs with 50 reference pairs.}
\label{fig:overview}
\end{figure*}

For an outcome comparison, we then ask how to estimate the difference using a limited labeling budget. A calibration counterexample shows why correcting the two scores separately is insufficient. We estimate the difference directly, using both agents' reference labels on sampled tasks and judge outputs on the full cohort. Regression fitted on the same 50 labeled pairs improves precision over direct paired sampling across the studied execution systems. Figure~\ref{fig:overview} connects the empirical problem, the evidence needed to inspect a comparison, and this practical estimation response.

\section{Related Work}
\paragraph{Evaluating web agents.}
AgentRewardBench checks model judgments against expert assessments and reports variation in agent scores and rankings across evaluators~\cite{Lu2025}. WebJudge improves evaluation by identifying task requirements and relevant screenshots~\cite{Xue2025}. Gaming the Judge shows that changing an agent's stated reasoning can distort judgments even when its actions and observations are fixed~\cite{Khalifa2026}. We fix the entire supplied record, including reasoning, and vary the judge's decision rule. Our outcome is the resulting change in the reported advantage between agents. Semantically equivalent instructions can also change judge scores~\cite{Bhat2026}. JudgeSense compares disagreement under reworded and repeated prompts~\cite{Bellibatlu2026}. Our repeated-prompt control addresses this variation while the main comparison changes the judgment target.

\paragraph{Outcomes and evidence.}
Judging agents can gather verification evidence by inspecting environments~\cite{Shi2026}. Sufficient-context analysis separates missing information from failures to use it~\cite{Joren2024}; abstention studies examine when models should withhold an answer or verdict~\cite{Kirichenko2025,Khoury2026}. In procedural-trace evaluation, linking evidence to the option it describes reduces false rejection~\cite{Chen2026Traces}. Our source experiments examine these relations in saved web-agent executions: we connect an observation to the relevant task condition and test how that changes the comparison.

\paragraph{Estimating outcomes from imperfect judgments.}
Binary error correction estimates outcome rates using a judge's sensitivity and specificity~\cite{Lee2025}. Conversational-score calibration aligns different judges to a shared human-derived scale~\cite{Zufle2026}. Prediction-powered inference uses predictions and labeled residuals; PPI++ learns a weight for the predictions~\cite{Angelopoulos2023,Angelopoulos2023PPIPlusPlus}. We examine how score correction affects the difference between agents, then assess direct estimation of that difference at a fixed paired-label budget.

\begin{table*}[t]
\centering
\caption{Judgments on the same 300 task pairs. P: prediction; Ct: certification. Rates are percentages; leads and lead reductions are percentage points. Each row reports a 95\% interval for the lead reduction.}
\label{tab:primary}
\begin{tabular}{llrrrrrrr}
\toprule
& & \multicolumn{2}{c}{Qwen rate} & \multicolumn{2}{c}{Claude rate} & \multicolumn{2}{c}{Claude lead} & Lead reduction \\
Judge & View & P & Ct & P & Ct & P & Ct & Estimate [95\% CI] \\
\midrule
Astra & Final & 18.33 & 13.17 & 26.67 & 14.50 & 8.33 & 1.33 & 7.00 [3.67, 10.50] \\
Astra & Expanded & 19.33 & 14.67 & 26.33 & 18.00 & 7.00 & 3.33 & 3.67 [0.50, 7.00] \\
Sonnet 4.5 & Final & 25.17 & 17.00 & 31.17 & 21.83 & 6.00 & 4.83 & 1.17 [$-$2.33, 4.83] \\
Sonnet 4.5 & Expanded & 23.33 & 17.17 & 30.50 & 21.00 & 7.17 & 3.83 & 3.33 [0.17, 6.67] \\
Kimi K2.5 & Final & 23.00 & 17.17 & 32.50 & 24.67 & 9.50 & 7.50 & 2.00 [$-$1.17, 5.33] \\
\bottomrule
\end{tabular}
\end{table*}

\section{Study Design}
\subsection{Fixed executions and judge inputs}
The main study uses all 300 shared test tasks for Qwen2.5-VL-72B and Claude 3.7 Sonnet in a fixed AgentRewardBench release. Tasks come from four benchmarks: AssistantBench, WebArena, VisualWebArena and WorkArena. A pair shares a task identifier, although generated entities and numerical instances can differ. Each execution is evaluated against its own recorded goal. We use the benchmark's development split separately for calibration and de\-vel\-op\-ment-fit controls.

Astra, Sonnet 4.5 and Kimi K2.5 serve as judges, distinct from the execution agents. All receive the task goal, full supplied action and reasoning trace, URLs, errors and referenced goal images. The \emph{final} view includes the terminal screenshot; the \emph{expanded} view adds up to two earlier screenshots. Astra and Sonnet receive both views, and Kimi receives the final view. The judge inputs exclude reference labels and environment rewards. Supplement U and the model-configuration record provide the exact model settings.

\subsection{Two questions about success}
\emph{Prediction} asks the judge to predict the execution's actual outcome. When the record does not settle completion, it must choose the more likely outcome. \emph{Certification} asks if the supplied record supports completion. Both use the same task-success criteria, evidence and binary response format; the changed instruction governs the decision under uncertainty. A certification zero means \emph{success not endorsed}, covering both observed failure and insufficient confirmation. We use ``endorsement rate'' when discussing these outputs.

Each execution receives two fresh calls per rule and view, whose endorsements are averaged. We report Claude's lead as its rate minus Qwen's rate, and the rule effect as the prediction lead minus the certification lead. A positive rule effect therefore means Claude's lead shrinks. The 95\% intervals use bootstrap resampling of task pairs within benchmarks, keeping each task's agents, rules and repetitions together. Full prompts, sampling settings and miss\-ing-verdict sensitivities are in the supplement.

\begin{table}[b]
\centering
\caption{Three-state judgments with Astra/final. Rates (\%) use the full denominator. Similar supported-success rates coexist with different failure and uncertainty rates.}
\label{tab:ternary}
\setlength{\tabcolsep}{4pt}
\begin{tabular}{lrrr}
\toprule
& Supported & Supported & \\
System & success & failure & Unresolved \\
\midrule
Qwen & 13.0 & 74.3 & 12.7 \\
Claude & 14.3 & 62.5 & 23.2 \\
\bottomrule
\end{tabular}
\end{table}

\subsection{Testing the reach of the comparison}
Two extensions address distinct questions. First, a three-state judge distinguishes supported success, supported failure and unresolved completion on the original executions. Fresh prediction and binary-certification controls use Astra/final in the same batch. This tests whether the lead reduction remains when uncertainty receives its own verdict. State rates use all scheduled judgments in their denominator.

Second, we include other execution systems from the same release. Qwen, Claude and GPT-4o share 300 tasks; adding Llama 3.3-70B leaves 206 common tasks because Llama has no VisualWebArena executions and lacks two WebArena tasks. Llama uses text observations, whereas the other released configurations use vision. All systems receive both rules with Astra and Sonnet on final-view records. Each comparison uses the full task intersection for its cohort. These overlapping cohorts broaden the systems examined while preserving the original study (Supplement S).

\section{The Rule Changes the Reported Advantage}
\subsection{The ranking can stay the same while the lead changes}
Table~\ref{tab:primary} shows why score changes must be interpreted jointly. In Astra/final, certification lowers Claude's endorsement rate more than Qwen's, reducing the lead from 8.33 to 1.33 points. The executions have not changed, and Claude still ranks first by the point estimates. What changes is the amount of improvement a reader would attribute to choosing Claude. The same observation has a different practical meaning from a uniform downward adjustment to both scores.

The effect varies across judge and view. All five point-estimate leads shrink, with the largest reduction in Astra/final; Sonnet/final and Kimi leave more uncertainty about the change. The ranking alone misses this variation. A fresh wording control also retains the Astra/final reduction under a rewrite of the instructions, supporting its interpretation as a consequence of the judgment question rather than a peculiarity of one phrasing. Complete wording and missing-output checks are provided in the supplement.

\begin{figure*}[tp]
\centering
\includegraphics[width=\textwidth]{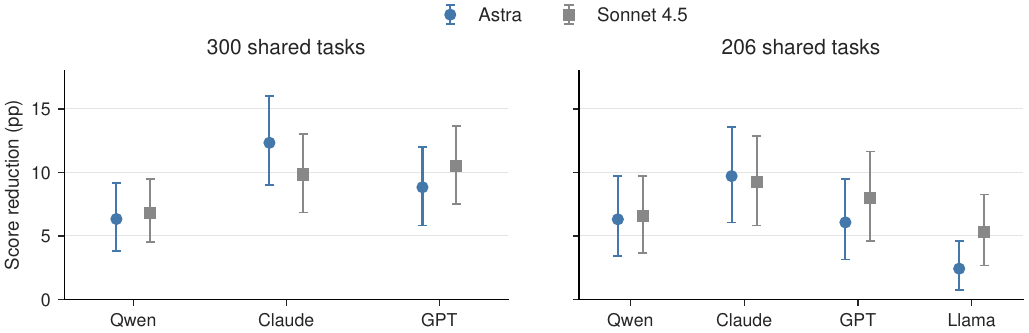}
\caption{Certification lowers the systems' scores by different amounts. Each point is a system's prediction-minus-certification rate, with a 95\% interval. The difference between two systems' reductions is the change in their lead. GPT denotes GPT-4o; the two task cohorts overlap.}
\Description{Two panels show rule-dependent score reductions for Qwen, Claude and GPT on 300 tasks and for those systems plus Llama on 206 tasks. Astra and Sonnet produce different patterns of reductions.}
\label{fig:extension}
\end{figure*}

\subsection{Explicit uncertainty preserves the effect}
Allowing an unresolved verdict preserves the central finding. In the fresh Astra/final batch, Claude's prediction lead of 8.17 points falls to a supported-success lead of 1.33 points under the three-state rule. The reduction is 6.83 points, with a 95\% interval of [3.33, 10.50]. The fresh binary-certification control also shows a reduction. Thus the effect remains when the judge can distinguish insufficient evidence from evidence of failure.

The state breakdown adds information that a binary rate conceals (Table~\ref{tab:ternary}). Claude has fewer supported failures but more unresolved judgments than Qwen. A small gap in supported success therefore does not mean the executions have the same outcome structure. Reporting uncertainty separately helps a reader distinguish a demonstrated failure from a record that does not settle the task. Supplement T gives the controls, transitions and complete state results.

\subsection{Different systems receive different reductions}
The additional execution systems also show unequal reductions (Figure~\ref{fig:extension}). Under Astra, the leads of Qwen, Claude and GPT over Llama all shrink on their common tasks. The pattern is less pronounced with Sonnet, and a few pairwise absolute gaps increase. These results broaden the evidence that a judgment rule can change the reported size of an improvement, while showing why no single correction to all agent scores would capture its effect. The full pairwise comparisons appear in Supplement S.

\section{What Evidence Makes a Comparison Checkable?}
\subsection{Recover relations needed by the task}
When a task requires a specific item, configuration or field, the evidence must connect that requirement to what the agent actually did. An observation can be relevant without making this connection: a product page may identify the purchased item but omit where it appeared in the requested grid. For a disputed comparison, the source question is which relation supports or challenges the task condition.

We examined archived records for every execution contributing to the original Astra/final change. Two human annotators checked the source-recovery annotations; a third resolved disagreements. Relevant relations could be located for 52 of the 59 executions, including records contributing in both directions. These include observations that support completion and ones that challenge it, identifying where a comparison can be checked against additional archived evidence. Supplement T contains the source census.

Both intervention studies below use Astra with the original final-view inputs and two repetitions per condition. We append archived observations while preserving the goal, action and reasoning trace, errors and original media. Re-presentation and partial-source controls test whether the relevant relation changes the judgment beyond repeating existing information. The complete intervention designs and results are in the supplement.

\begin{table}[htbp]
\centering
\caption{Source additions change local comparisons. Entries are (Qwen, Claude) success endorsements under prediction (P) and certification (Ct); both repetitions agree. Full controls are in the supplement.}
\label{tab:sources}
\begin{tabular}{@{}p{0.66\columnwidth}cc@{}}
\toprule
Evidence supplied & P & Ct \\
\midrule
\multicolumn{3}{@{}l}{\textit{Grid position}} \\
Original record & (1,1) & (0,0) \\
Add position--product--destination relation & (0,1) & (0,1) \\
\addlinespace
\multicolumn{3}{@{}l}{\textit{Laptop purchase}} \\
Record + button mapping & (0,1) & (0,0) \\
Add product identity & (1,1) & (0,0) \\
Add product identity and configuration & (1,1) & (1,1) \\
\bottomrule
\end{tabular}
\end{table}

\subsection{A missed distinction in a positional task}
In VisualWebArena task 756 (Figure~\ref{fig:overview}(b), Case 1), the goal is to select a product in the second row and second column. Qwen clicks the second item in the first row; Claude clicks the requested position and reaches the corresponding product page. The original inputs fail to preserve the relation between grid position, product and destination. Prediction endorses both executions, while certification endorses neither. Both rules therefore report a zero gap despite the different actions.

Adding the relations among positions, products and destination pages makes both rules distinguish the executions (Table~\ref{tab:sources}). Re-presenting existing observations does not produce this change. The earlier screenshots alone also leave the requested off-screen position unclear; saved layout information supplies the relevant relation. The useful addition is the connection needed to check the instruction, rather than simply another view of the final product page.

\subsection{An apparent advantage in a purchase task}
A laptop purchase gives the reverse example (Figure~\ref{fig:overview}(b), Case 2). The task requests a Development Laptop (PC), while the confirmation names a Dell XPS 13. With the recorded click target mapped to the \emph{Order Now} button, prediction interprets Qwen's execution as a wrong purchase and endorses only Claude. The archived product page connects these two names; supplying that relation resolves the mistaken identity and removes the lead. Configuration evidence alone does not resolve the identity question; certification endorses both executions only when both identity and configuration are supplied. The two rules can thus retain the same zero gap while changing their judgments of both agents.

Together, these cases show what a source intervention should accomplish: make the task condition checkable against the relevant object and observation. Its value is not determined by whether the lead grows or shrinks. Supplying the relation also does not ensure that every judge will use it correctly. Supplementary query and form experiments retain examples of both improved use and persistent object/time confusion. For practice, the immediate recommendation is to preserve the links needed to inspect a disputed judgment, rather than treat a longer record as sufficient evidence by itself.

\section{Estimating the Intended Agent Difference}
\subsection{Better scores can give a worse difference}
An outcome comparison needs an explicit reference. The human reference uses AgentRewardBench's released expert assessments of task completion; the functional reference uses its saved benchmark rewards converted to binary outcomes. They imply substantially different Claude--Qwen leads: 10.67 points for the human reference and 0.67 for the functional reference. Each fixes what the estimator should recover before any calibration is fitted. A researcher must therefore choose the completion criterion that matches the intended claim, alongside deciding what evidence the judge receives.

Separate score calibration does not necessarily improve that target difference. A judged gap's error is the difference between the two agents' signed rate errors. If both rates are underestimated by similar amounts, those errors largely cancel. Reducing each absolute error can disrupt that cancellation, leaving a worse difference. This is why a close judged gap can be misleading evidence that the individual scores are well calibrated, and why individual calibration improvements do not settle the comparison.

We observe this failure using binary error correction fitted separately for each agent on development labels~\cite{Lee2025}. Table~\ref{tab:calibration} shows the Astra/final certification example against the functional reference. Initially, both scores underestimate their targets by similar amounts. After calibration, Qwen is slightly below its target and Claude is above it. Each score is closer, but their errors now pull the estimated lead in opposite directions. Other settings include joint improvements; the practical lesson is to check whether calibration improves the claimed difference. Supplement F.10 provides the complete calibration results.

\begin{table}[htbp]
\centering
\caption{Closer scores, a worse lead estimate. Signed errors for Astra/final certification against the functional reference (percentage points); positive means overestimation.}
\label{tab:calibration}
\begin{tabular}{lrr}
\toprule
Estimated quantity & Before calibration & After calibration \\
\midrule
Qwen rate & $-6.50$ & $-1.52$ \\
Claude rate & $-5.83$ & $+5.37$ \\
\midrule
Claude--Qwen lead & $+0.67$ & $+6.89$ \\
\bottomrule
\end{tabular}
\end{table}

\begin{figure*}[tp]
\centering
\includegraphics[width=\textwidth]{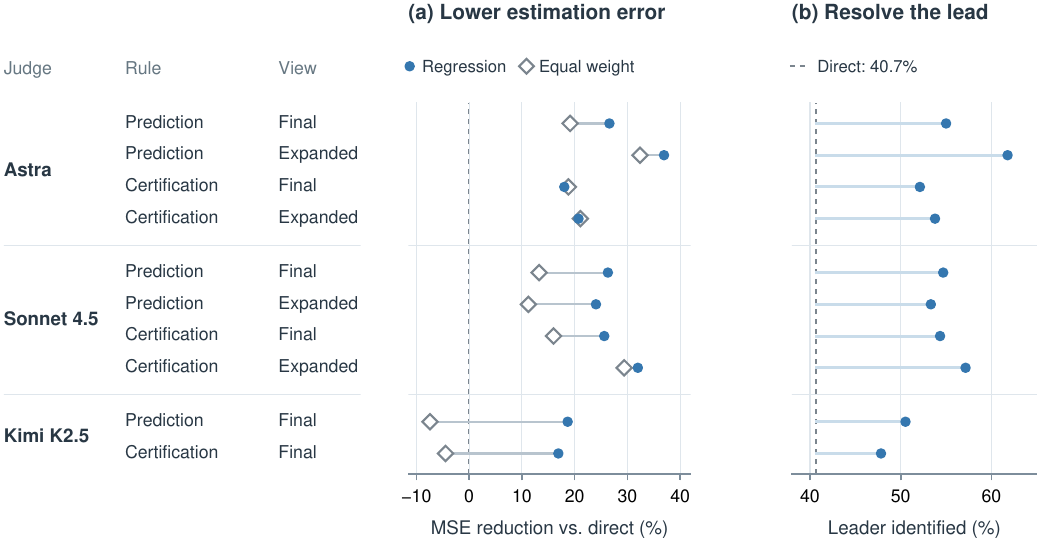}
\caption{More informative comparisons from the same 50 labeled task pairs, using the human reference. \textbf{(a)} MSE reductions relative to direct paired sampling: blue circles show regression and gray diamonds show equal-weight correction. Positive values indicate lower error. \textbf{(b)} How often regression's approximate 95\% interval identifies the reference leader; the dashed line marks direct sampling. Full numerical results and coverage are in Supplement V.}
\Description{Ten settings grouped by judge share row labels. The left panel compares regression and equal-weight MSE reductions against direct paired sampling. The right panel shows regression intervals identifying the reference leader, with direct sampling as a common baseline.}
\label{fig:estimation}
\end{figure*}

\subsection{Use paired labels to estimate the gap directly}
Existing judgments can still help estimate the chosen difference even when their raw gap is inaccurate. Suppose we can label both agents' executions for a random sample of tasks. That sample gives a direct estimate of the reference gap. Labeling both executions for each sampled task keeps the two sides of the comparison aligned. The sample may nevertheless overrepresent tasks on which one agent tends to do well. Comparing the sample's judge gap with the full cohort's judge gap provides information about this sampling imbalance. A correction uses that information to reduce estimation error, as illustrated in Figure~\ref{fig:overview}(c).

Let $Y_i$ denote the agents' reference outcome difference on task $i$, and $Z_i$ the corresponding difference in judge endorsements. For a random sample $S$ of task pairs, we estimate the reference difference as~\cite{Angelopoulos2023}
\begin{equation}
\widehat\mu=\bar Y_S+\hat{\beta}(\bar Z-\bar Z_S),
\label{eq:estimator}
\end{equation}
where $\bar Z$ is the judge difference over the full cohort. The first term is the labeled sample's reference gap. The second adjusts it using the difference between full-cohort and sampled judgments. The fitted weight $\hat{\beta}$ controls how much of this discrepancy to use. The baselines use fixed weights: $\beta=0$ for direct paired sampling and $\beta=1$ for equal-weight correction.

We learn the auxiliary weight from the labeled pairs~\cite{Angelopoulos2023PPIPlusPlus}, regressing the paired reference differences on the paired judge differences within the same 50 labeled task pairs. This reuses the labels needed for the comparison instead of requiring a separate development set. All baselines use the same paired labels, so the comparison isolates the benefit of auxiliary judgments. The supplement retains fixed-weight, de\-vel\-op\-ment-fit and label-splitting controls.

The judge's useful contribution is information about variation across tasks. A systematic offset in its reported gap cancels in $\bar Z-\bar Z_S$, whereas a useful association between $Z_i$ and $Y_i$ can reduce sampling error. This explains why neither a small raw gap error nor a large response to the judgment rule is required for an auxiliary judge to help. Supplement S.6 gives a runnable estimator using reference labels only for the sampled pairs.

\subsection{More precise comparisons with the same labels}
We evaluate the full procedure by repeatedly sampling 50 task pairs without replacement from the fixed cohort. Every regression is fitted only on the sampled labels; the known full-cohort reference is used to measure its error. The methods share the same samples, and mean squared error (MSE) counts both variance and squared bias. Fifty task pairs require labels for 100 executions. Complete sampling and interval methods appear in Supplements F.11 and U.

Against the human reference, regression reduces MSE by 16.9--36.9\% relative to direct paired sampling across all ten settings (Figure~\ref{fig:estimation}). Equal weighting slightly outperforms regression in the two Astra certification settings. Learning how much to trust a judge is useful in many settings, while retaining the fixed-weight control reveals where the additional fit does little.

The precision gain also helps answer a practical comparison question. Using the same approximate 95\% interval procedure, regression identifies the human-reference leader in about 48--62\% of samples, compared with 41\% for direct sampling. Its measured coverage is 94.2--94.7\%, close to direct sampling's 94.6\%. Thus the intervals more often resolve the ordering at the same label budget, with coverage slightly below the nominal level. Exact decision frequencies, wrong-direction declarations and coverage are reported together in Supplement V.

Kimi illustrates the distinction between rule sensitivity and estimation value. Its rule contrast is comparatively small and uncertain, yet its judgments improve regression precision over both direct sampling and equal weighting. The study of additional execution systems extends the precision result to every included pair under both the human and functional references. Regression improves on direct paired sampling throughout; equal weighting remains competitive in a few human-reference settings. Supplement S reports all settings and the exceptions.

\section{Implications for Agent Evaluation}
\subsection{What should an improvement claim report?}
The first decision is what the claim means. A likely-outcome estimate and an evidence-supported success rate serve different purposes. Naming that target makes comparisons easier to interpret, and reporting both agents' rates alongside the gap shows whether a rule affects them equally. For an evidence-support target, separating supported failure from unresolved completion makes the result more informative than a binary non-success rate.

The next decision concerns a disputed judgment. Preserve the source relation needed to check the task: which item occupied the requested position, which configuration was submitted, or which query produced a count. Our source examples show why this can change the comparison in either direction. They also provide a concrete question to ask of a record, beyond whether it contains enough screenshots or a confident completion claim.

When the objective is an outcome difference, paired reference labels provide a direct way to estimate it. With full-cohort judgments available, randomly sample tasks, label both agents on each, and use the auxiliary differences to improve precision. Only the sampled executions need reference labels; the remaining tasks contribute judge predictions. Same-label regression makes this feasible without purchasing a separate development set. The benefit concerns reference-label use; acquiring the full set of judge outputs is a separate cost. Researchers should assess both estimation error and interval coverage for the intended use.

\subsection{Scope}
The study uses fixed released executions, and the source interventions establish local effects of particular evidence relations. The estimation results measure sampling risk within the declared cohorts and references. New tasks, execution interfaces or judge versions may change these results. The complete supplement preserves the comparison matrices, source controls and alternative estimators needed to assess that scope.

\balance
\section{Conclusion}
The size of an agent's reported advantage deserves its own evaluation. A stable ranking can conceal a substantially changed lead, and more accurate individual scores can still produce a less accurate difference. Our experiments connect these problems to the success target and its evidence, then show how paired regression can improve estimation with limited reference labels. This gives researchers a practical way to support claims about how much an agent has improved.

\bibliographystyle{ACM-Reference-Format}
\bibliography{references}
\end{document}